\documentclass[11pt]{article}

\usepackage[preprint]{acl}

\usepackage{times}
\usepackage{latexsym}
\usepackage{hyperref}

\usepackage[T1]{fontenc}

\usepackage[utf8]{inputenc}

\usepackage{microtype}

\usepackage{inconsolata}

\usepackage{graphicx}
\usepackage{amsmath}
\usepackage{amssymb}
\usepackage{algorithm}
\usepackage{algorithmic}
\usepackage{booktabs}
\usepackage{makecell}
\usepackage{multirow}
\usepackage{float}
\usepackage[most]{tcolorbox}
\newcommand{\methodname}{CLAD}
\newtcolorbox{insightbox}{
    colback=gray!8,
    colframe=gray!45,
    boxrule=0.5pt,
    arc=2pt,
    left=6pt,
    right=6pt,
    top=5pt,
    bottom=5pt,
    width=\linewidth
}

\usepackage{array}
\usepackage{listings}

\lstdefinestyle{appendixcode}{
    language=Python,
    basicstyle=\ttfamily\scriptsize,
    breaklines=true,
    breakatwhitespace=false,
    columns=flexible,
    keepspaces=true,
    showstringspaces=false,
    aboveskip=0pt,
    belowskip=0pt,
    xleftmargin=0pt,
    xrightmargin=0pt
}

\usepackage{array}
\usepackage{fvextra}

\newcommand{\rowhead}[1]{\textbf{\mbox{#1}}}

\DefineVerbatimEnvironment{CodeBlock}{Verbatim}{
    fontsize=\small,
    breaklines=true,
    breakanywhere=true,
    tabsize=4,
    baselinestretch=1.0
}

\usepackage{xcolor}

\newcommand{\accup}[2]{\ensuremath{#1_{{\color{green!50!black}\uparrow #2}}}}
\newcommand{\accdown}[2]{\ensuremath{#1_{{\color{red!70!black}\downarrow #2}}}}
\newcommand{\accsame}[1]{\ensuremath{#1_{{\color{gray}\scriptstyle\,0.00}}}}

\usepackage[normalem]{ulem}

\title{Cluster-Level Attention-Guided Parallel Decoding for Masked Diffusion Language Models}

\author{
  \textbf{Heqiang Qi}\textsuperscript{1},
  \textbf{Wei Huang}\textsuperscript{2,3},
  \textbf{Mingyuan Bai}\textsuperscript{4},
  \textbf{Xiangming Meng}\textsuperscript{1}\thanks{Corresponding author.} \\
  \textsuperscript{1}Zhejiang University \quad
  \textsuperscript{2}RIKEN Center for Advanced Intelligence Project \\
  \textsuperscript{3}The Institute of Statistical Mathematics \\
  \textsuperscript{4}Agency for Science, Technology and Research (A$\star$STAR) \\
  \texttt{heqiangqi@zju.edu.cn},
  \texttt{wei.huang.vr@riken.jp} \\
  \texttt{yvonne.mingyuanbai@outlook.com},
  \texttt{xiangmingmeng@intl.zju.edu.cn}
}

\begin{document}
\maketitle
\begin{abstract}

Masked diffusion language models (MDLMs) enable parallel decoding by predicting all masked positions at each denoising step, yet existing training-free samplers usually decide which positions to commit at token-level granularity. We revisit this granularity and observe that reliable predictions often emerge as contiguous high-confidence spans, suggesting that the unit of parallel commitment can be larger than a single token. We first group adjacent high-confidence candidates into confidence-induced clusters (CICs) as span-level update units. We then use self-attention maps from the same forward pass to estimate inter-cluster dependencies, enabling conflict-aware selection of mutually compatible CICs for parallel commitment. This yields \methodname{} (\textbf{C}luster-\textbf{L}evel \textbf{A}ttention-Guided \textbf{D}ecoding), a training-free cluster-level decoder for MDLMs. Experiments on LLaDA and Dream model families across four reasoning and code-generation benchmarks show that \methodname{} achieves $1.77\times$--$8.47\times$ speedups over Vanilla decoding while maintaining broadly comparable task accuracy in most settings. \footnote{Code is released at \url{https://github.com/ziyigu2004/CLAD}.}

\end{abstract}

\section{Introduction}
\label{sec:introduction}

Masked diffusion language models (MDLMs) ~\citep{sahoo2024simple,shi2024simplified,ou2025your,nie2026large,ye2025dream} have recently emerged as a promising alternative to autoregressive language models ~\citep{brown2020language,ouyang2022training, achiam2023gpt,bai2023qwen,grattafiori2024llama,guo2025deepseek}  for text generation. Instead of generating tokens strictly from left to right, MDLMs start from a masked sequence and iteratively recover masked positions. Since all currently masked positions can be predicted within each denoising step, the decoding policy must decide which predictions should be committed and used as context for subsequent steps.

\begin{figure}[t]
    \centering
    \makebox[\linewidth][c]{%
        \includegraphics[
            width=1.2\linewidth,
            trim=140 50 140 50,
            clip
        ]{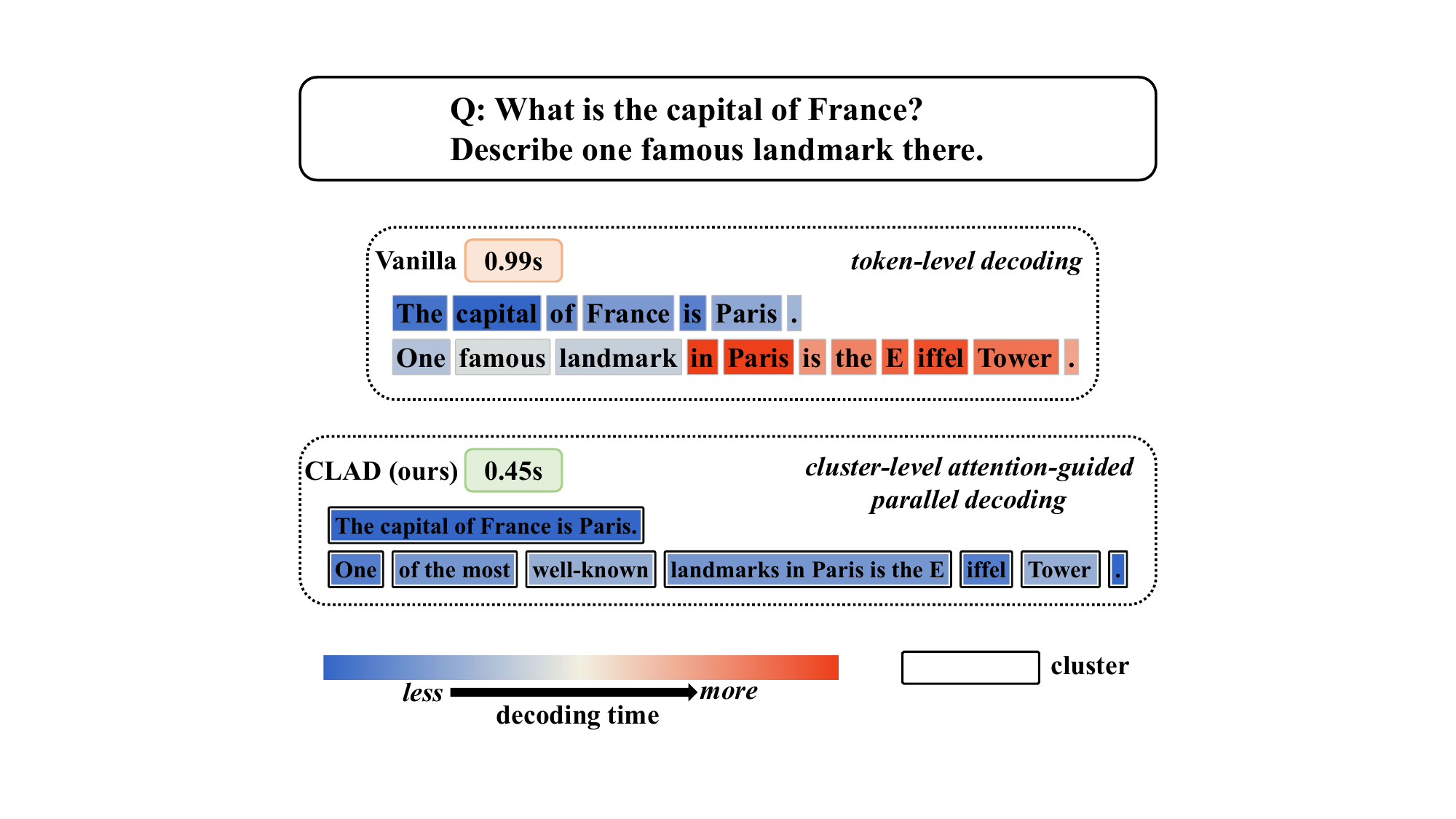}
    }
    \caption{
    Qualitative comparison of Vanilla decoding and \methodname{}.
    Colors indicate relative decoding time, with cooler colors denoting earlier commitments and warmer colors denoting later commitments; tokens with the same color are committed in parallel.
    Unlike Vanilla token-level decoding, \methodname{} commits contiguous high-confidence spans as cluster-level update units, reaching the same correct answer with lower latency.
    }
    \label{fig:introduction}
\end{figure}

This commitment decision largely determines how much parallelism an MDLM decoder can exploit without excessive quality degradation ~\citep{feng2026theoretical,kang2025parallelbench}.  A decoder that commits more reliable predictions per step can reduce the number of denoising iterations and improve throughput. However, overly aggressive commitment may accept predictions that are reliable in isolation yet incompatible when committed together. Existing training-free samplers mostly make this decision at token-level granularity. Confidence- and uncertainty-based methods rank individual masked positions by maximum probability, entropy, margin, or distributional stability ~\citep{wu2025fast,kim2025train,ben2026accelerated,kim2026klass}, while dependency-aware methods estimate token-level interactions from self-attention to avoid committing strongly coupled positions ~\citep{kim2026dependency,luo2026dawn}. Despite these advances, the basic scheduling unit remains an individual token position.

We revisit this decoding granularity. Although MDLMs expose token-wise predictive distributions, token positions need not be the only useful unit of decoding progress. This view is consistent with recent efforts that decouple global semantic organization from local textual realization in language diffusion models ~\citep{guo2026continuous}, as well as observations that local neighborhoods around high-confidence predictions can become reliable during MDLM decoding ~\citep{kong2025accelerating,du2026r}. This leads to the central question of this work:
\emph{Can an MDLM decoder commit contiguous high-confidence spans as update units while avoiding strongly dependent spans in the same denoising step?}

The first part of this question concerns the unit of progress. We observe that high-confidence candidates often emerge as contiguous spans rather than isolated positions. Once such a span becomes well supported by the current partially unmasked context, committing its tokens one by one can spend unnecessary refinement steps on a local region that is already predictable. We therefore define a \emph{confidence-induced cluster} (CIC) as a maximal contiguous span of high-confidence masked positions. A CIC serves as an operational notion of a locally reliable span: its positions are adjacent, individually confident, and can be treated as a candidate update unit.

The second part concerns compatibility. Cluster-level commitment should not simply decode all confident spans at once. Even if each CIC is locally reliable, different CICs may remain mutually dependent under the current context. Committing such clusters simultaneously may accept predictions that are confident marginally but inconsistent jointly. We extend this attention-based dependency signal from individual positions to CICs, using self-attention maps from the same forward pass to estimate inter-cluster dependencies and model them as \emph{dependency conflicts}. This yields a simple cluster-level view of MDLM decoding: CICs define the units of progress, while attention-derived conflicts determine which units are compatible for same-step commitment.

Based on this view, we propose \methodname{} (\textbf{C}luster-\textbf{L}evel \textbf{A}ttention-Guided \textbf{D}ecoding), a training-free cluster-level decoder for MDLMs. At each denoising step, \methodname{} converts token-level confidence estimates into CIC-level update candidates, constructs a sparse conflict graph over CICs using attention-derived inter-cluster dependencies, and selects a maximum-weight set of non-conflicting CICs for parallel commitment. By keeping only mutually strongest inter-cluster dependencies, \methodname{} obtains a compact conflict graph and performs cluster selection efficiently.

We evaluate \methodname{} on mathematical reasoning and code-generation benchmarks using pretrained MDLMs from the LLaDA and Dream families ~\citep{nie2026large,zhu2025llada,ye2025dream}. Across four benchmarks, \methodname{} achieves $1.77\times$--$8.47\times$ speedups over Vanilla decoding while maintaining broadly comparable task accuracy in most settings. Compared with token-level dependency-aware baselines such as DAPD ~\citep{kim2026dependency} and DAWN ~\citep{luo2026dawn}, \methodname{} improves throughput in most evaluated settings. These results suggest that changing the decoding unit from tokens to confidence-induced spans is an effective way to improve throughput, while attention-derived compatibility modeling helps preserve accuracy under more aggressive parallel commitment.

Our contributions are summarized as follows:
\begin{itemize}
    \item We revisit the decoding granularity of MDLMs and propose confidence-induced clusters (CICs), maximal contiguous spans of high-confidence masked positions, as explicit span-level update units.

    \item We propose \methodname{} (\textbf{C}luster-\textbf{L}evel \textbf{A}ttention-Guided \textbf{D}ecoding), a training-free cluster-level decoder that builds attention-derived inter-cluster conflict graphs and selects compatible CICs for parallel commitment.

    \item Experiments on four representative reasoning and code-generation benchmarks across the LLaDA and Dream model families show that  \methodname{} achieves $1.77\times$--$8.47\times$ speedups over Vanilla decoding while maintaining broadly comparable task accuracy in most settings. 
\end{itemize}

\section{Related Work}
\label{sec:related_works}

\paragraph{Discrete Diffusion Language Models.}
The pioneering work D3PM ~\citep{austin2021structured} establishes the paradigm for early discrete diffusion models, which typically define a corruption process over categorical variables through transition matrices. Subsequent work further generalizes this line of research with continuous-time Markov chain formulations ~\citep{campbell2022continuous} and score-based discrete diffusion models ~\citep{meng2022concrete}. More recently, masked diffusion language models have attracted increasing attention due to their simple absorbing-state corruption process and strong empirical performance ~\citep{sahoo2024simple,shi2024simplified,ou2025your}.
The scaling of masked diffusion language models to large language model regimes has been actively pursued. LLaDA ~\citep{nie2026large} demonstrates competitive performance on reasoning, instruction following, and code generation tasks. Dream ~\citep{ye2025dream} further explores diffusion language modeling at scale by leveraging autoregressive initialization.

\paragraph{Decoding Strategies for MDLMs.}
Existing training-free samplers for MDLMs mostly make the commitment decision at token-level granularity. Confidence-based decoding ranks individual masked positions by the maximum probability of their predicted token distributions ~\citep{chang2022maskgit,nie2026large}. Other uncertainty-aware criteria have also been explored, including entropy-based selection ~\citep{koh2024plm} and margin-based confidence ~\citep{kim2025train}. To improve inference efficiency, recent methods commit multiple token positions in parallel while attempting to control quality degradation. Fast-dLLM ~\citep{wu2025fast} accelerates inference by combining confidence-threshold parallel decoding with cache optimization. EB-Sampler ~\citep{ben2026accelerated} selects tokens under an entropy-bounded criterion, while KLASS ~\citep{kim2026klass} uses distributional stability across denoising steps to identify reliable tokens. Dependency-aware methods further refine this token-level commitment decision with self-attention ~\citep{vaswani2017attention} signals. DOS ~\citep{zhou2026dependency} and Attn-Sampler ~\citep{zhou2026attention} prioritize masked tokens according to their dependence on the current unmasked context, while DAWN ~\citep{luo2026dawn} and DAPD ~\citep{kim2026dependency} use attention-derived token interactions to avoid simultaneously decoding strongly coupled masked positions. 

A related line of work exploits local structure beyond isolated token
positions. R$^2$-dLLM ~\citep{du2026r} reduces decoding redundancy
via local confidence aggregation and stability checks, while
LocalLeap ~\citep{kong2025accelerating} relaxes parallel decoding around locally deterministic high-confidence anchors. 

\section{Preliminaries}
\label{sec:preliminaries}

\paragraph{Masked Diffusion Language Models.}
Masked diffusion language models formulate text generation as the
reverse of an absorbing-state masking process. Let
$\mathbf{x}_0=(x_0^1,\ldots,x_0^L)\in\mathcal{V}^L$ denote a clean token
sequence, where $\mathcal{V}$ is the vocabulary and $L$ is the sequence
length. The vocabulary is augmented with a special mask token
$m=\mathtt{[MASK]}$. 

The forward corruption process is governed by a
monotonically decreasing schedule $\alpha_t$, with $\alpha_0\approx 1$
and $\alpha_1\approx 0$. At timestep $t\in[0,1]$, each token is
independently preserved with probability $\alpha_t$ or replaced by the
absorbing mask state:
\begin{equation}
q(\mathbf{x}_t\mid\mathbf{x}_0) = \prod_{i=1}^{L} \mathrm{Cat}\!\left(x_t^i;\, \alpha_t\delta_{x_0^i} + (1-\alpha_t)\delta_m \right).
\label{eq:forward}
\end{equation}
As $t$ increases, $\alpha_t$ decreases and the sequence is progressively
corrupted toward the fully masked state.

The generative process reverses this corruption by iteratively
unmasking tokens. Given a denoising model $f_\theta$, the reverse
transition from timestep $t$ to an earlier timestep $s$ is parameterized
as
\begin{equation}
\small
p_\theta(x_s^i\mid\mathbf{x}_t) =
\left\{
\begin{array}{@{}l@{\qquad}l@{}}
\mathrm{Cat}\!\left(x_s^i;\delta_{x_t^i}\right),
& x_t^i\neq m, \\[5pt]
\multicolumn{2}{@{}l@{}}{
\mathrm{Cat}\!\left(
x_s^i;\,
\frac{1-\alpha_s}{1-\alpha_t}\delta_m
+
\frac{\alpha_s-\alpha_t}{1-\alpha_t} f_\theta^i(\mathbf{x}_t)
\right),
} \\[-1pt]
\multicolumn{2}{@{}r@{}}{x_t^i = m .}
\end{array}
\right.
\label{eq:reverse}
\end{equation}
Here $f_\theta^i(\mathbf{x}_t)$ denotes the predicted distribution over
clean vocabulary tokens at position $i$. Therefore, already visible
tokens are deterministically carried over, while masked tokens are
gradually converted into clean tokens according to the model prediction.

The denoising model is trained by predicting the original tokens at
masked positions, yielding the following training objective:
\begin{equation}
\mathcal{L}(\theta) = \mathrm{E}_{\mathbf{x}_0,t,\mathbf{x}_t} \left[
\frac{\alpha_t'}{1-\alpha_t} \sum_{i:x_t^i=m} \log f_\theta^i(x_0^i\mid\mathbf{x}_t)
\right].
\label{eq:objective}
\end{equation}
After training, generation starts from a masked sequence and repeatedly
applies the reverse process. Since all masked positions can be predicted
simultaneously at each denoising step, the decoding strategy that
selects which positions to commit becomes a key factor for both
generation accuracy and inference efficiency.

\section{Methodology}
\label{sec:Methodology}

\subsection{Overview}
\label{sec:motivation}

\begin{figure*}[htbp]       
    \centering
    \includegraphics[width=1.0\linewidth]{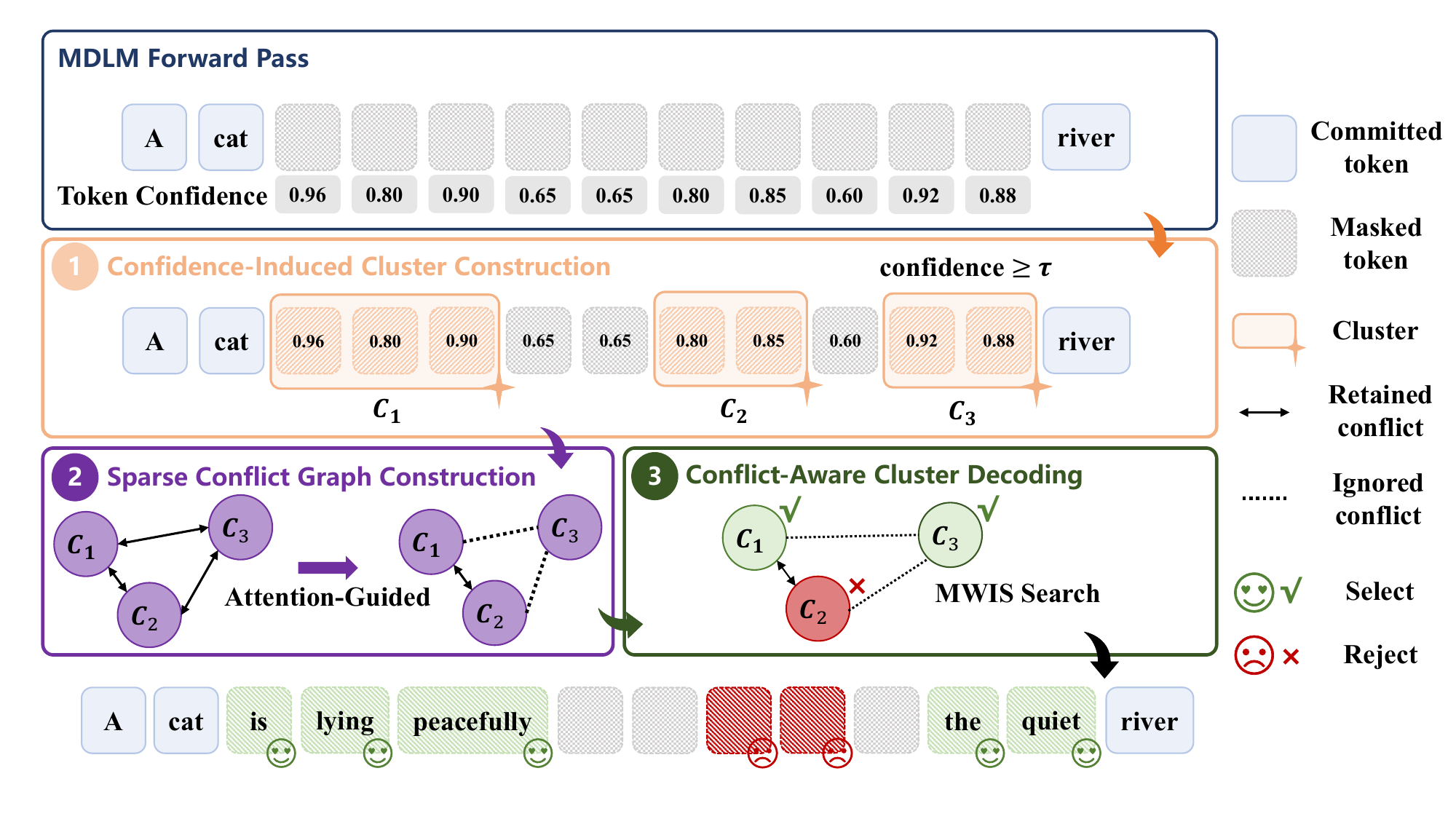}
    \caption{Overview of \methodname{}. The framework contains three main components: Confidence-Induced Cluster Construction, Sparse Conflict Graph Construction, and Conflict-Aware Cluster Decoding.}
    \label{fig:overview}
\end{figure*}

The introduction motivates a cluster-level view of MDLM decoding: the decoder can improve throughput by committing locally reliable spans rather than isolated token positions, but it must avoid committing span-level units that remain strongly dependent in the same denoising step. We now instantiate this view as \methodname{}, a training-free parallel decoder that operates directly on the model's current predictions and attention maps.

Figure~\ref{fig:overview} summarizes the overall procedure. At each denoising step, \methodname{} uses the model's token-level confidence estimates to construct \emph{confidence-induced clusters} (CICs), i.e., maximal contiguous spans of high-confidence masked positions. These CICs serve as span-level update candidates and are the main source of acceleration: committing a selected CIC reveals multiple adjacent tokens in one step rather than spending separate refinement steps on each token.

However, faster commitment should not mean committing every confident span. Different CICs may still depend on each other under the current partially unmasked context, so committing them simultaneously can introduce incompatible updates. We refer to this incompatibility as a \emph{dependency conflict}. To mitigate such conflicts, \methodname{} reuses self-attention maps from the same forward pass to estimate inter-cluster dependencies and constructs a sparse conflict graph. Nodes in this graph are CICs, and edges indicate pairs of CICs that should not be committed together. The final decoding step selects a set of non-conflicting CICs and commits their greedy predictions in parallel, thereby increasing the number of tokens committed per step while avoiding cluster pairs estimated to be strongly dependent.

\begin{figure}[H]
    \centering
    \setlength{\tabcolsep}{1pt}
    \begin{tabular}{@{}c@{\hspace{0.35em}}c@{}}
        \includegraphics[
            height=0.45\columnwidth,
            trim=130pt 39pt 138pt 24pt,
            clip
        ]{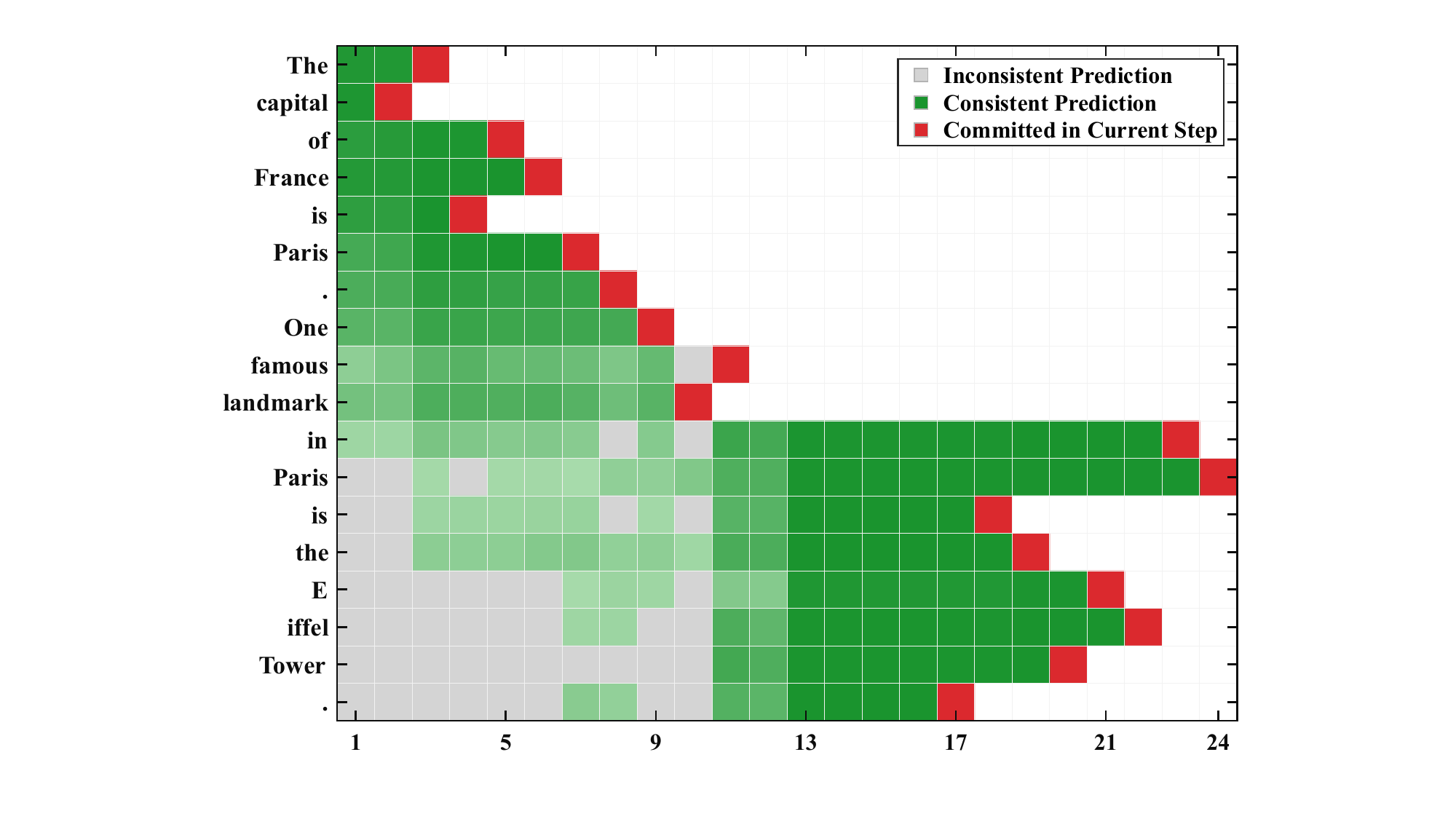}
        &
        \includegraphics[
            height=0.45\columnwidth,
            trim=226pt 13pt 400pt 20pt,
            clip
        ]{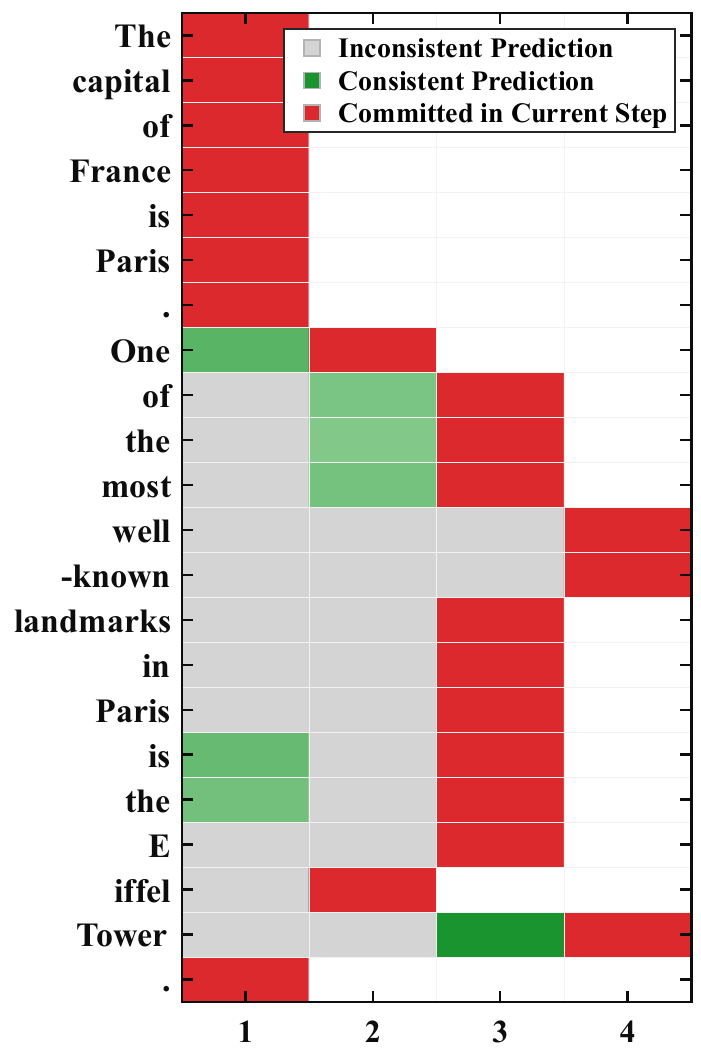}
        \\[-0.2em]
        {\scriptsize\textbf{(a) Vanilla}} &
        {\scriptsize\textbf{(b) CLAD }}
    \end{tabular}

    \caption{
    Detailed decoding traces for the two decoding processes illustrated in Figure~\ref{fig:introduction}. (a) shows the Vanilla process, while (b) shows the CLAD process. Green, gray, and red cells denote consistent predictions, inconsistent predictions, and tokens committed at the current denoising step, respectively.
    }
    \label{fig:motivation}
\end{figure}

Figure~\ref{fig:motivation} provides a detailed trace of this behavior. The example illustrates the primary acceleration mechanism of \methodname{}: once several adjacent positions become reliable together, CLAD can commit them as a single cluster-level update. Conflict-aware selection is applied when multiple candidate CICs are present, deciding which span-level updates can be safely committed in the same step. The following subsections formalize the three components of \methodname{}: confidence-induced cluster construction, sparse conflict graph construction, and conflict-aware cluster decoding.

\subsection{Confidence-Induced Cluster Construction}
\label{sec:confidence_Induced_cluster}

At denoising step $t$, let $\mathbf{x}_t$ denote the current sequence state and $\mathcal{M}_t=\{i:x_t^i=m\}$ denote the set of masked positions. For each masked position, the model produces a predictive distribution. We define the greedy prediction and confidence as
\begin{align}
\hat{x}_0^{i} &= \arg\max_{v\in\mathcal{V}} p_\theta(x_0^i = v\mid\mathbf{x}_t), \label{eq:argmax} \\
c^{i} &= p_\theta(x_0^i=\hat{x}_0^i\mid\mathbf{x}_t). \label{eq:confidence}
\end{align}
A position is regarded as a candidate if its confidence is above a threshold $\tau$:
\begin{equation}
\mathcal{P}_t
=
\{i\in\mathcal{M}_t:c^i\geq\tau\}.
\label{eq:candidate_set}
\end{equation}

The candidate set is then decomposed into maximal contiguous spans. We denote the resulting span set as
\begin{equation}
\mathcal{C}_t=\{C_1,\ldots,C_K\},
\quad
C_k=\{l_k,l_k+1,\ldots,r_k\}.
\label{eq:cluster_set}
\end{equation}
Each $C_k$ is a \emph{confidence-induced cluster (CIC)}. By construction, all positions inside $C_k$ are high-confidence candidates, and no adjacent candidate outside $[l_k,r_k]$ can be added to the cluster. 

\subsection{Sparse Conflict Graph Construction}
\label{sec:cluster_graph}


CICs provide span-level candidates for faster commitment, but confidence alone does not determine whether different CICs can be safely committed together. Strongly dependent CICs may produce incompatible same-step updates, so we treat salient inter-cluster dependency as a \emph{dependency conflict}. To identify these conflicts, we estimate inter-cluster dependency using the attention map produced by the same forward pass. The resulting conflict graph connects pairs of CICs that should not be committed in the same denoising step.

Let $A_t\in\mathbb{R}^{L\times L}$ be the head-averaged attention matrix at step $t$. Since the graph is used to detect conflicts rather than directional generation dependencies, we symmetrize the attention scores:
\begin{equation}
R^{ij}_{t} = \max\!\left(A^{ij}_{t},A^{ji}_{t}\right), \quad R^{ii}_{t}=0.
\label{eq:sym_attention}
\end{equation}

For two clusters $C_a$ and $C_b$, their dependency score is computed by aggregating token-level attention into a cluster-level interaction:
\begin{equation}
\small
s_{ab}
=
\max
\left\{
\max_{i\in C_a}
\frac{1}{|C_b|}
\sum_{j\in C_b}R^{ij}_{t},
\;
\max_{j\in C_b}
\frac{1}{|C_a|}
\sum_{i\in C_a}R^{ji}_{t}
\right\}.
\label{eq:cluster_score}
\end{equation}
This score measures the strongest cross-cluster dependency in either direction. The averaging operation avoids favoring longer target clusters, while the maximum operation captures the most influential token-level interaction between two clusters.

We then construct a sparse conflict graph $G_{t}=(\mathcal{C}_{t},\mathcal{E}_{t})$, where the node set $\mathcal{C}_t$ consists of all CICs at step $t$, and the edge set $\mathcal{E}_t$ records inter-cluster conflicts. Instead of connecting all pairs with large attention scores, we retain only reciprocal salient dependencies. For each cluster $C_a$, let
\begin{equation}
b^\star(a)=\arg\max_{b\neq a}s_{ab}
\label{eq:top_cluster}
\end{equation}
be its strongest external dependency. An edge is added between $C_a$ and $C_b$ when they are mutually strongest dependencies and their interaction is above the average dependency level of both clusters:
\begin{equation}
(a,b)\in\mathcal{E}_{t}
\Longleftrightarrow
\left\{
\begin{array}{l}
b=b^\star(a),\quad a=b^\star(b),\\[2pt]
s_{ab}\geq \bar{s}_a,\quad s_{ab}\geq \bar{s}_b,
\end{array}
\right.
\label{eq:conflict_edge}
\end{equation}
where
\begin{equation}
\bar{s}_a=
\frac{1}{K-1}\sum_{b\neq a}s_{ab},\quad K\neq 1.
\label{eq:mean_score}
\end{equation}
When $K=1$, we skip graph construction and directly select the only CIC.
This mutual-top criterion retains only salient reciprocal conflicts. The resulting sparse graph directly indicates which CICs should not be committed together, without introducing an additional conflict-threshold hyperparameter. Edges are selected adaptively based on reciprocal salience and the cluster-specific average scores \(\bar{s}_a\) and \(\bar{s}_b\).

\subsection{Conflict-Aware Cluster Decoding}
\label{sec:cluster_scheduling}

Given the conflict graph, the goal is to commit as many reliable tokens as possible while avoiding conflicting clusters. We formulate this as a maximum-weight independent set problem (MWIS) ~\citep{nemhauser1975vertex}:
\begin{equation}
\begin{aligned}
\mathcal{S}_{t}
&=
\arg\max_{\mathcal{S}\subseteq\mathcal{C}_{t}}
\sum_{C_a\in\mathcal{S}} |C_a|, \\
&\quad
\mathrm{s.t.}\ 
(C_a,C_b)\notin\mathcal{E}_{t},
\quad \forall C_a,C_b\in\mathcal{S}.
\end{aligned}
\label{eq:mwis}
\end{equation}
Although MWIS is generally hard on arbitrary graphs, our mutual-top construction makes the resulting conflict graph a matching, where each CIC participates in at most one retained conflict edge.
Therefore, Eq.~\eqref{eq:mwis} can be solved exactly rather than greedily: we select all isolated CICs and, for each retained conflict edge, keep the CIC that contains more token positions, i.e., the higher-weight CIC.
This selection costs \(O(K+|\mathcal{E}_t|)\) after graph construction, where \(K=|\mathcal{C}_t|\).
The selected CICs are therefore mutually non-conflicting, and the final update set is the union of the selected clusters:
\begin{equation}
\mathcal{U}_{t} = \bigcup_{C_a\in\mathcal{S}_{t}} C_a.
\label{eq:update_set}
\end{equation}
The sequence is updated by committing the greedy prediction at every
selected position:
\begin{equation}
x^i_{t-1}
=
\left\{
\begin{array}{ll}
\hat{x}^i_{0}, & i\in\mathcal{U}_{t},\\[2pt]
x^i_{t}, & i\notin\mathcal{U}_{t}.
\end{array}
\right.
\label{eq:update_rule}
\end{equation}

If no CIC is formed because no masked position passes the confidence threshold, we fall back to committing the single masked position with the highest confidence, ensuring that the denoising process always makes progress.

\begin{algorithm}[t]
\small
\caption{\methodname{}: Cluster-Level Attention-Guided Decoding}
\label{alg:clad}
\begin{algorithmic}[1]
\REQUIRE Current state $\mathbf{x}_{t}$, model $p_\theta$, confidence threshold $\tau$
\ENSURE Updated state $\mathbf{x}_{t-1}$

\STATE $\mathcal{M}_{t}\leftarrow\{i:x_t^i=m\}$
\IF{$\mathcal{M}_{t}=\emptyset$}
    \RETURN $\mathbf{x}_{t}$
\ENDIF

\STATE $(\hat{\mathbf{x}}_{0}, \mathbf{c}, A_t)
\leftarrow \mathrm{Forward}(p_\theta,\mathbf{x}_t)$
\STATE $\mathcal{P}_{t}\leftarrow\{i\in\mathcal{M}_{t}:c^i\geq\tau\}$

\IF{$\mathcal{P}_{t}=\emptyset$}
    \STATE $j^\star\leftarrow \arg\max_{i\in\mathcal{M}_{t}} c^i$
    \STATE $\mathbf{x}_{t-1}\leftarrow\mathbf{x}_{t}$
    \STATE $x_{t-1}^{j^\star}\leftarrow \hat{x}_{0}^{j^\star}$
    \RETURN $\mathbf{x}_{t-1}$
\ENDIF

\STATE $\mathcal{C}_{t}\leftarrow \mathrm{CIC\  Construction}(\mathcal{P}_{t})$
\STATE $G_t=(\mathcal{C}_t,\mathcal{E}_t)
\leftarrow \mathrm{Sparse\ Graph}(\mathcal{C}_{t},A_t)$

\STATE $\mathcal{S}_{t}\leftarrow
\mathrm{MWIS}(G_t,\omega(C_a)=|C_a|)$

\STATE $\mathcal{U}_{t}\leftarrow
\bigcup_{C_a\in\mathcal{S}_{t}}C_a$
\STATE $\mathbf{x}_{t-1}\leftarrow\mathbf{x}_{t}$
\STATE $x_{t-1}^{i}\leftarrow \hat{x}_{0}^{i},\quad \forall i\in\mathcal{U}_{t}$

\RETURN $\mathbf{x}_{t-1}$
\end{algorithmic}
\end{algorithm}

The above components together define the decoding procedure of
\methodname{}. 
Algorithm~\ref{alg:clad} summarizes the overall procedure. 

\section{Experiments}
\label{sec:experiments}


\subsection{Experimental Setup}
\label{sec:experimental_setup}

\begin{table*}[!t]
\centering
\small
\setlength{\tabcolsep}{5pt}
\caption{
Main results on reasoning and code generation benchmarks across LLaDA-8B-Instruct and Dream-v0-Instruct-7B. Acc. denotes task accuracy, TPS denotes tokens per second, and Speedup is measured relative to Vanilla top-1 decoding under the same setting. Colored subscripts on Acc. report the absolute change relative to Vanilla. Higher values are better for all metrics, and the best TPS and Speedup are highlighted in bold.
}
\label{tab:main_results}
\resizebox{\textwidth}{!}{
\begin{tabular}{llccc ccc}
\toprule
\multirow{2}{*}{Task} 
& \multirow{2}{*}{Method}
& \multicolumn{3}{c}{LLaDA-8B-Instruct}
& \multicolumn{3}{c}{Dream-v0-Instruct-7B} \\
\cmidrule(lr){3-5} \cmidrule(lr){6-8}
& & Acc.$\uparrow$ & TPS$\uparrow$ & Speedup$\uparrow$
  & Acc.$\uparrow$ & TPS$\uparrow$ & Speedup$\uparrow$ \\
\midrule

\multirow{6}{*}{GSM8K (5-shot)}
& Vanilla    & 77.71 & 5.22  & $1.00\times$ & 76.49 & 3.30  & $1.00\times$ \\
& Fast-dLLM  & \accup{78.32}{0.61} & 15.92 & $3.05\times$ & \accdown{75.89}{0.60} & 13.27 & $4.02\times$ \\
& KLASS      & \accdown{75.59}{2.12} & 12.40 & $2.38\times$ & \accdown{73.46}{3.03} & 5.40  & $1.64\times$ \\
& DAPD       & \accdown{74.07}{3.64} & 14.64 & $2.80\times$ & \accdown{70.51}{5.98} & 7.95 & $2.41\times$ \\
& DAWN       & \accdown{76.88}{0.83} & 21.94 & $4.20\times$ & \accdown{74.37}{2.12} & \textbf{15.64} & $\mathbf{4.74\times}$ \\
& \methodname{} (ours) 
              & \accup{77.79}{0.08} & \textbf{25.59} & $\mathbf{4.90\times}$ & \accdown{73.69}{2.80} & 14.54 & $4.41\times$ \\
\midrule

\multirow{6}{*}{MATH (4-shot)}
& Vanilla    & 32.84 & 8.25  & $1.00\times$ & 38.36 & 7.88  & $1.00\times$ \\
& Fast-dLLM  & \accdown{32.80}{0.04} & 21.82 & $2.64\times$ & \accdown{38.12}{0.24} & 16.95 & $2.15\times$ \\
& KLASS      & \accdown{31.96}{0.88} & 15.62 & $1.89\times$ & \accdown{34.78}{3.58} & 7.72  & $0.98\times$ \\
& DAPD       & \accdown{31.44}{1.40} & 21.08 & $2.56\times$ & \accdown{32.00}{6.36} & 13.11 & $1.66\times$ \\
& DAWN       & \accdown{32.36}{0.48} & 28.76 & $3.49\times$ & \accdown{37.98}{0.38} & 19.69 & $2.50\times$ \\
& \methodname{} (ours)
              & \accdown{31.92}{0.92} & \textbf{31.01} & $\mathbf{3.76\times}$ & \accdown{37.34}{1.02} & \textbf{20.44} & $\mathbf{2.59\times}$ \\
\midrule

\multirow{6}{*}{MBPP (3-shot)}
& Vanilla    & 31.60 & 5.45  & $1.00\times$ & 54.80 & 3.74  & $1.00\times$ \\
& Fast-dLLM  & \accup{31.80}{0.20} & 20.01 & $3.67\times$ & \accup{56.00}{1.20} & 17.71 & $4.74\times$ \\
& KLASS      & \accup{31.96}{0.36} & 15.62 & $2.87\times$ & \accup{56.20}{1.40} & 9.44  & $2.52\times$ \\
& DAPD       & \accup{32.00}{0.40} & 8.08 & $1.48\times$ & \accdown{54.40}{0.40} & 5.35 & $1.43\times$ \\
& DAWN       & \accdown{30.60}{1.00} & 26.53 & $4.87\times$ & \accup{56.00}{1.20} & 20.15 & $5.39\times$ \\
& \methodname{} (ours)
              & \accdown{30.20}{1.40} & \textbf{26.66} & $\mathbf{4.89\times}$ & \accdown{54.00}{0.80} & \textbf{20.64} & $\mathbf{5.52\times}$ \\
\midrule

\multirow{6}{*}{HumanEval (0-shot)}
& Vanilla    & 40.24 & 16.66 & $1.00\times$ & 52.44 & 11.07 & $1.00\times$ \\
& Fast-dLLM  & \accup{41.46}{1.22} & 55.11 & $3.31\times$ & \accup{56.71}{4.27} & 26.80 & $2.42\times$ \\
& KLASS      & \accsame{40.24} & 31.81 & $1.91\times$ & \accup{54.27}{1.83} & 12.80 & $1.16\times$ \\
& DAPD       & \accdown{39.63}{0.61} & 32.51 & $1.95\times$ & \accdown{51.83}{0.61} & 12.28 & $1.11\times$ \\
& DAWN       & \accsame{40.24} & 67.99 & $4.08\times$ & \accup{54.27}{1.83} & 31.37 & $2.83\times$ \\
& \methodname{} (ours)
              & \accsame{40.24} & \textbf{70.65} & $\mathbf{4.24\times}$ & \accup{54.27}{1.83} & \textbf{31.62} & $\mathbf{2.86\times}$ \\
\bottomrule
\end{tabular}
}
\end{table*}

\paragraph{Models.}
We evaluate \methodname{} on four open-source masked diffusion language models, including LLaDA-8B-Instruct ~\citep{nie2026large}, LLaDA-1.5 ~\citep{zhu2025llada}, Dream-v0-Base-7B, and Dream-v0-Instruct-7B ~\citep{ye2025dream}. We keep model weights fixed and only modify the decoding strategy, so all compared methods are evaluated in a training-free setting.

\paragraph{Datasets.}
We evaluate on benchmarks covering mathematical reasoning and code generation. For mathematical reasoning, we use GSM8K (5-shot) ~\citep{cobbe2021training} and MATH (4-shot) ~\citep{hendrycks2021measuring}, which require multi-step reasoning and longer-form answer generation. For code generation, we use MBPP (3-shot) ~\citep{austin2021program} and HumanEval (0-shot) ~\citep{chen2021evaluating}, which require generating executable Python solutions from natural language specifications.

\paragraph{Baselines.}
We compare \methodname{} with the following training-free decoding methods. \textbf{Vanilla} denotes the original top-1 decoding strategy of MDLMs. \textbf{Fast-dLLM} ~\citep{wu2025fast} dynamically commits positions whose confidence exceeds a predefined threshold. \textbf{KLASS} ~\citep{kim2026klass} selects tokens using both confidence and token-level distributional stability across denoising steps. \textbf{DAPD} ~\citep{kim2026dependency} and \textbf{DAWN} ~\citep{luo2026dawn} use attention-derived token-level dependency graphs to guide dependency-aware parallel commitment. These baselines represent both uncertainty-based and dependency-aware training-free decoding strategies.

\paragraph{Evaluation Metrics.}
We report accuracy and tokens per second (TPS). For code generation, accuracy is measured by the pass rate on the official unit tests. For mathematical reasoning, accuracy is computed by matching the extracted final answer against the ground-truth answer. TPS is measured end-to-end over the evaluation generation loop, including prompt preprocessing, decoding, and output post-processing.

\paragraph{Implementation Details.}
Unless otherwise specified, we use the same generation length 256 and block length 32 for all compared methods under each benchmark. 
Additional implementation details are provided in Appendix~\ref{app:implementation_details}.

\subsection{Main Results}
\label{sec:Main Results}

Table~\ref{tab:main_results} reports the accuracy and decoding speed of different methods across two masked diffusion language models and four benchmarks. Overall, \methodname{} substantially improves decoding throughput in most settings while maintaining broadly comparable task accuracy. On LLaDA-8B-Instruct, \methodname{} achieves the highest TPS on all four benchmarks, obtaining $4.90\times$, $3.76\times$, $4.89\times$, and $4.24\times$ speedup on GSM8K, MATH, MBPP, and HumanEval, respectively. On Dream-v0-Instruct-7B, \methodname{} also achieves the best throughput on three out of four benchmarks, including MATH, MBPP, and HumanEval. At the same time, we acknowledge that \methodname{} still shows modest accuracy drops in a few settings, particularly on GSM8K with Dream-v0-Instruct-7B ($-2.80$ points) and MBPP with LLaDA-8B-Instruct ($-1.40$ points). Additional results on LLaDA-1.5 and Dream-v0-Base-7B are in Appendix~\ref{app:additional_results}.

Compared with DAWN and DAPD, which perform dependency-aware scheduling at the token level, \methodname{} achieves higher throughput in most settings. This supports the central motivation of \methodname{}: using CICs as scheduling units can increase effective parallelism by committing multiple adjacent high-confidence candidates in a single update. Meanwhile, the broadly comparable accuracy suggests that conflict-aware selection helps maintain generation quality under this more aggressive cluster-level commitment strategy. Overall, \methodname{} provides a more attractive alternative to other token-level parallel decoding methods.


\subsection{Ablation Study}
\label{sec:ablation}

\begin{figure}[H]
    \centering
    \begin{minipage}[t]{0.5\linewidth}
        \centering
        \includegraphics[
            width=\linewidth,
            trim=140 300 140 300,
            clip
        ]{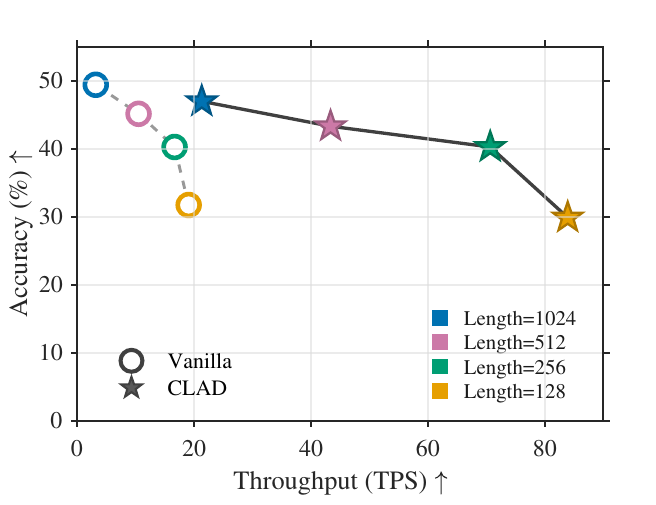}
        \vspace{-0.6em}
        \centerline{\scriptsize (a) LLaDA-8B-Instruct}
    \end{minipage}%
    \hfill
    \begin{minipage}[t]{0.5\linewidth}
        \centering
        \includegraphics[
            width=\linewidth,
            trim=140 300 140 300,
            clip
        ]{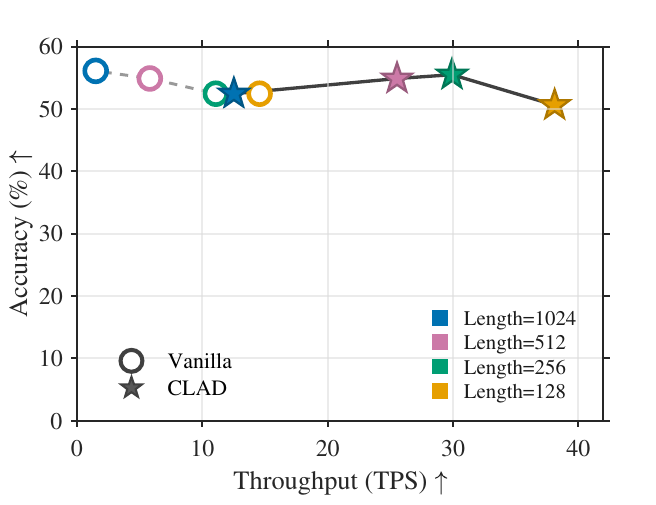}
        \vspace{-0.6em}
        \centerline{\scriptsize (b) Dream-v0-Instruct-7B}
    \end{minipage}

    \caption{
    Ablation study on generation length using HumanEval (0-shot).
    Circles denote Vanilla decoding, and stars denote
    \methodname{}; colors indicate different generation lengths.
    The x-axis reports throughput in tokens per second (TPS), and the
    y-axis reports accuracy.
    }
    \label{fig:length_ablation}
\end{figure}

\paragraph{Impact of generation length.}
We study the effect of generation length on HumanEval (0-shot) by varying the generation budget in $\{128,256,512,1024\}$. 
As shown in Figure~\ref{fig:length_ablation}, \methodname{} achieves higher throughput than Vanilla decoding across all lengths while maintaining broadly comparable accuracy, with the largest gains at medium lengths. 
This indicates that cluster-level commitment remains effective under different generation budgets.

\begin{figure}[H]
    \centering
    \begin{minipage}[t]{0.48\linewidth}
        \centering
        \includegraphics[
            width=\linewidth,
            trim=180 320 180 320,
            clip]{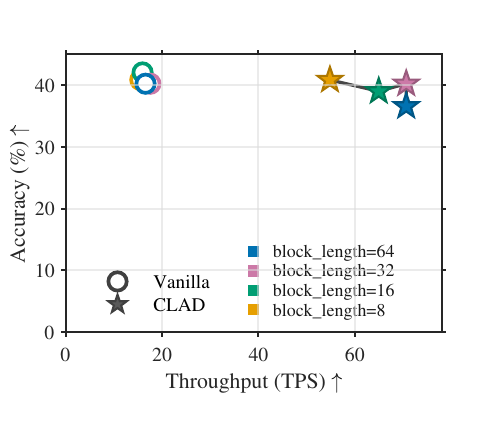}
        \vspace{-0.6em}
        \centerline{\scriptsize (a) LLaDA-8B-Instruct}
    \end{minipage}
    \hfill
    \begin{minipage}[t]{0.48\linewidth}
        \centering
        \includegraphics[
            width=\linewidth,
            trim=180 320 180 320,
            clip]{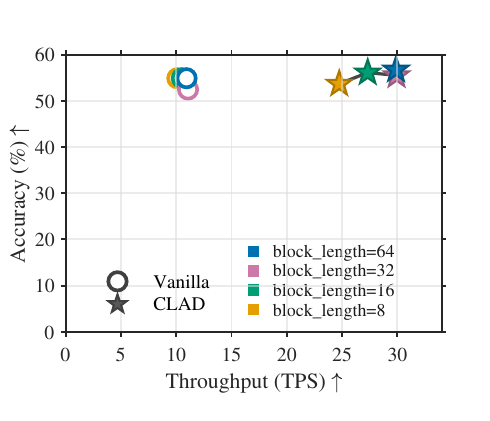}
        \vspace{-0.6em}
        \centerline{\scriptsize (b) Dream-v0-Instruct-7B}
    \end{minipage}
    \caption{
    Ablation study on block size using HumanEval (0-shot). Circles denote Vanilla decoding, and stars denote \methodname{}; colors indicate different block
    sizes. The x-axis reports throughput in tokens per second (TPS), and
    the y-axis reports accuracy.
    }
    \label{fig:block_ablation}
\end{figure}

\paragraph{Impact of block size.}
We next examine the effect of block size with the generation length fixed to 256. Figure~\ref{fig:block_ablation} shows that \methodname{} achieves higher throughput under all block sizes on both models. For LLaDA, \methodname{} achieves its highest speedup at a block size of 32, while increasing the block size further preserves high throughput with a slight accuracy decrease. On Dream, \methodname{} remains robust across block sizes and improves both throughput and accuracy in most settings. 

\begin{figure}[t]
    \centering
    \begin{minipage}[t]{0.48\linewidth}
        \centering
        \includegraphics[
            width=\linewidth,
            trim=180 330 180 330,
            clip]{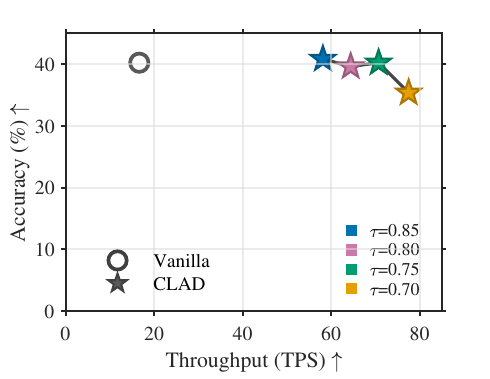}
        \vspace{-0.6em}
        \centerline{\scriptsize (a) LLaDA-8B-Instruct}
    \end{minipage}
    \hfill
    \begin{minipage}[t]{0.48\linewidth}
        \centering
        \includegraphics[
            width=\linewidth,
            trim=180 330 180 330,
            clip]{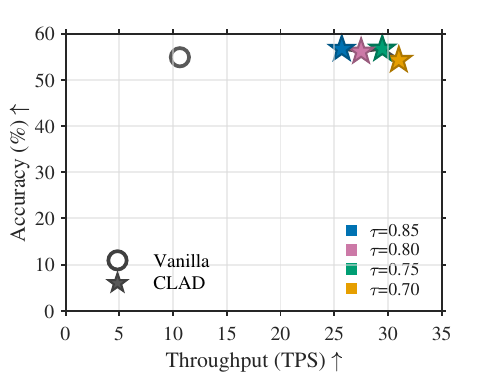}
        \vspace{-0.6em}
        \centerline{\scriptsize (b) Dream-v0-Instruct-7B}
    \end{minipage}
    \caption{
    Ablation study on the confidence threshold $\tau$ using HumanEval
    (0-shot). Circles denote Vanilla decoding, and stars denote \methodname{} with
    different thresholds. Colors indicate different values of $\tau$.
    The x-axis reports throughput in tokens per second (TPS), and the
    y-axis reports accuracy.
    }
    \label{fig:tau_ablation}
\end{figure}

\paragraph{Impact of confidence threshold $\tau$.}
We also analyze the confidence threshold $\tau$, which controls the candidate positions used to form CICs. As shown in Figure~\ref{fig:tau_ablation}, smaller thresholds admit more candidates and therefore improve throughput, but may also introduce less reliable updates. Larger thresholds make candidate selection more selective, which tends to stabilize accuracy at the cost of reduced speed. Across both models, moderate thresholds generally provide a practical balance between throughput and accuracy.


\paragraph{Component ablation.}
Table~\ref{tab:ablation_components} shows that CIC alone brings high speed but degrades accuracy.
Randomly selecting clusters after constructing the conflict graph still fails to recover accuracy.
With MWIS, \methodname{} matches Vanilla accuracy while reaching 70.65 TPS, validating the necessity of conflict graph-based cluster selection.

\begin{table}[t]
\centering
\scriptsize
\setlength{\tabcolsep}{3pt}
\caption{
Component ablation on HumanEval (0-shot) with LLaDA-8B-Instruct.
Graph denotes the conflict graph.
}
\label{tab:ablation_components}
\resizebox{\columnwidth}{!}{
\begin{tabular}{lccccc}
\toprule
Variant & CIC & Graph & MWIS & Acc. & TPS \\
\midrule
Vanilla & $\times$ & $\times$ & $\times$ & 40.24 & 16.66 \\
+CIC & $\checkmark$ & $\times$ & $\times$ & 36.58 & 76.03 \\
+Graph & $\checkmark$ & $\checkmark$ & $\times$ & 36.58 & 66.13 \\
+MWIS & $\checkmark$ & $\checkmark$ & $\checkmark$ & 40.24 & 70.65 \\
\bottomrule
\end{tabular}
}
\end{table}

\section{Conclusion}
\label{sec:conclusion}

We proposed \methodname{}, a training-free cluster-level parallel decoder for MDLMs. The key idea is to move beyond token-level commitment by grouping adjacent high-confidence candidates into confidence-induced clusters and using them as span-level update units. To control the risk of aggressive cluster-level commitment, \methodname{} estimates inter-cluster dependencies from self-attention and selects a set of non-conflicting CICs for parallel decoding. 
Experiments on LLaDA and Dream model families show that \methodname{} substantially improves decoding throughput while maintaining broadly comparable task accuracy in most settings.


\section*{Limitations}
\label{sec:limitations}

\methodname{} uses self-attention as a lightweight proxy for estimating inter-cluster dependencies when constructing the conflict graph. This choice is straightforward, efficient, and intuitive: attention maps are already available from the same forward pass, require no additional training, and provide a useful signal for identifying clusters that may interact during same-step commitment. Empirically, this signal works well for guiding conflict-aware selection. However, attention is still an indirect heuristic rather than a theoretically grounded quantitative measure of causal dependence. A large attention score does not necessarily imply that two clusters will lead to incompatible updates, and a small score does not fully rule out latent semantic or logical dependencies. Therefore, the conflict graph should be viewed as an efficient approximation for reducing risky parallel commitments, rather than a formal guarantee of cluster independence.

The benefit of \methodname{} may also vary across models and tasks. Cluster-level decoding is most effective when the current denoising state contains locally coherent spans that can be committed together with limited inter-cluster incompatibility. When candidate updates are highly fragmented, when dependencies among clusters are dense, or when a task requires more fine-grained step-by-step refinement, the available parallelism can be smaller. Different MDLM backbones may also exhibit different attention patterns and decoding dynamics, which can affect how reliably CICs serve as update units. Thus, \methodname{} should not be interpreted as universally optimal for every setting; rather, it provides a practical cluster-level decoding strategy whose effectiveness depends on how well the model and task support reliable span-level commitment.

\bibliography{custom}

\clearpage
\appendix

\section{Additional Implementation Details}
\label{app:implementation_details}

Experiments on LLaDA-8B-Instruct, LLaDA-1.5, and Dream-v0-Base-7B are conducted on a single NVIDIA RTX 4090 GPU, while experiments on Dream-v0-Instruct-7B are conducted on a single NVIDIA L40 GPU. For a fair comparison, the attention signal used by DAPD, DAWN, and \methodname{} is computed in the same way, by averaging attention maps over all heads from the last four Transformer layers.

\section{Parameter Configurations}
\label{app:params}

\newcommand{\param}[1]{\texttt{#1}}

For compactness, we use LLaDA-Inst to denote LLaDA-8B-Instruct, Dream-Base to denote Dream-v0-Base-7B, and Dream-Inst to denote Dream-v0-Instruct-7B. We use the public implementations of Fast-dLLM, KLASS, and DAWN with their default or author-recommended hyperparameters. Since DAPD does not provide an official open-source implementation, we re-implement it following the original paper. For both DAPD and CLAD, we perform a small preliminary sweep over their method-specific thresholds on a limited subset of examples, and use the selected configurations for the full evaluation. The sweep is guided by how much parallelism an MDLM decoder can exploit without excessive quality degradation. All selected hyperparameter settings are provided below.

\paragraph{(a) Fast-dLLM.}
Fast-dLLM ~\citep{wu2025fast} is an uncertainty-based parallel decoding method. Its main hyperparameter is \param{threshold}. In our experiments, Fast-dLLM uses a fixed confidence threshold of 0.9 for all evaluated models and tasks.

\begin{table}[H]
\centering
\footnotesize
\setlength{\tabcolsep}{3.5pt}
\renewcommand{\arraystretch}{1.08}
\caption{
KLASS hyperparameters. The two LLaDA models use the same settings, and
the two Dream models use the same settings. HE denotes HumanEval.
}
\label{tab:klass_params}
\begin{tabular}{@{}llcccc@{}}
\toprule
Model & Param. & GSM8K & MATH & HE & MBPP \\
\midrule
\multirow{2}{*}{LLaDA models}
& \(\tau_{\mathrm{conf}}\) & 0.6 & 0.6 & 0.9 & 0.7 \\
& \(\tau_{\mathrm{KL}}\)   & 0.015 & 0.01 & 0.01 & 0.01 \\
\midrule
\multirow{2}{*}{Dream models}
& \(\tau_{\mathrm{conf}}\) & 0.9 & 0.9 & 0.8 & 0.9 \\
& \(\tau_{\mathrm{KL}}\)   & 0.001 & 0.005 & 0.001 & 0.001 \\
\bottomrule
\end{tabular}
\end{table}

\paragraph{(b) KLASS.}
KLASS ~\citep{kim2026klass} selects candidate tokens using both prediction confidence and distributional stability across denoising steps. We denote its confidence threshold by \(\tau_{\mathrm{conf}}\) and its KL-stability threshold by \(\tau_{\mathrm{KL}}\). The confidence threshold filters out low-confidence predictions, while the KL-stability threshold requires the predicted token distribution to remain stable between consecutive denoising steps; a smaller \(\tau_{\mathrm{KL}}\) imposes a stricter stability requirement. Table~\ref{tab:klass_params} reports the task-specific values of
\(\tau_{\mathrm{conf}}\) and \(\tau_{\mathrm{KL}}\).

\paragraph{(c) DAPD.}
DAPD ~\citep{kim2026dependency} uses attention-based dependency estimates to avoid committing
strongly dependent tokens in the same denoising step. Its main
hyperparameters are \(\tau_{\min}\) and \(\tau_{\max}\), which control the
lower and upper bounds of the dependency criterion. Other settings are
fixed across experiments: the switching ratio is \(0.5\), the fast-stage
confidence threshold is \(0.9\), and single-block decoding is disabled.
Table~\ref{tab:dapd_params} reports the task-specific values of
\(\tau_{\min}\) and \(\tau_{\max}\). 

\begin{table}[H]
\centering
\footnotesize
\setlength{\tabcolsep}{4pt}
\renewcommand{\arraystretch}{1.08}
\caption{
DAPD hyperparameters. HE denotes HumanEval.
}
\label{tab:dapd_params}
\begin{tabular}{@{}llcc@{}}
\toprule
Model & Param. & GSM8K / MATH & HE / MBPP \\
\midrule
LLaDA-1.5
& \(\tau_{\min}\) & 0.0005 & 0.0005 \\
& \(\tau_{\max}\) & 0.0020 & 0.0010 \\
\midrule
LLaDA-Inst
& \(\tau_{\min}\) & 0.0010 & 0.0005 \\
& \(\tau_{\max}\) & 0.0040 & 0.0012 \\
\midrule
Dream-Base
& \(\tau_{\min}\) & 0.0005 & 0.0004 \\
& \(\tau_{\max}\) & 0.0020 & 0.0008 \\
\midrule
Dream-Inst
& \(\tau_{\min}\) & 0.0010 & 0.0005 \\
& \(\tau_{\max}\) & 0.0040 & 0.0010 \\
\bottomrule
\end{tabular}
\end{table}

\paragraph{(d) DAWN.}
DAWN ~\citep{luo2026dawn} constructs a token-level dependency graph from attention maps and uses this graph to guide parallel commitment. Its hyperparameters control different parts of the graph-based decoder. The threshold \(\tau_{\mathrm{sink}}\) filters attention-sink positions, \(\tau_{\mathrm{edge}}\) controls whether an attention link is retained as a dependency edge, \(\tau_{\mathrm{induce}}\) controls the confidence requirement for anchor-induced candidate tokens, and \(\tau_{\mathrm{conf}}\) controls which remaining candidate tokens are considered for conflict-based scheduling. Table~\ref{tab:dawn_fixed_params} and Table~\ref{tab:dawn_conf_params} report the fixed graph-construction hyperparameters and the task-specific \(\tau_{\mathrm{conf}}\), respectively.

\begin{table}[H]
\centering
\footnotesize
\setlength{\tabcolsep}{6pt}
\renewcommand{\arraystretch}{1.08}
\caption{
DAWN fixed graph-construction hyperparameters.
}
\label{tab:dawn_fixed_params}
\begin{tabular}{@{}lccc@{}}
\toprule
Model & \(\tau_{\mathrm{sink}}\) & \(\tau_{\mathrm{induce}}\) & \(\tau_{\mathrm{edge}}\) \\
\midrule
LLaDA-1.5  & 0.01 & 0.70 & 0.07 \\
LLaDA-Inst & 0.01 & 0.70 & 0.07 \\
Dream-Base & 0.03 & 0.75 & 0.05 \\
Dream-Inst & 0.03 & 0.75 & 0.10 \\
\bottomrule
\end{tabular}
\end{table}

\begin{table}[H]
\centering
\footnotesize
\setlength{\tabcolsep}{5pt}
\renewcommand{\arraystretch}{1.08}
\caption{
DAWN task-specific \(\tau_{\mathrm{conf}}\). HE denotes HumanEval.
}
\label{tab:dawn_conf_params}
\begin{tabular}{@{}lcccc@{}}
\toprule
Model & GSM8K & MATH & HE & MBPP \\
\midrule
LLaDA-1.5  & 0.75 & 0.75 & 0.80 & 0.75 \\
LLaDA-Inst & 0.75 & 0.75 & 0.80 & 0.70 \\
Dream-Base & 0.75 & 0.80 & 0.80 & 0.80 \\
Dream-Inst & 0.80 & 0.80 & 0.80 & 0.80 \\
\bottomrule
\end{tabular}
\end{table}

\paragraph{(e) \methodname{}.}
\methodname{} has a single tunable hyperparameter, the confidence
threshold \(\tau\), which determines whether a masked position is included
as a high-confidence candidate for CIC construction. A lower threshold
allows more positions to form CICs and leads to more aggressive parallel
commitment, while a higher threshold makes decoding more conservative.
Table~\ref{tab:clad_params} reports the task-specific values of \(\tau\).

\begin{table}[H]
\centering
\footnotesize
\setlength{\tabcolsep}{5pt}
\renewcommand{\arraystretch}{1.08}
\caption{
\methodname{} hyperparameters. Each entry reports the confidence threshold
\(\tau\) used for CIC construction. HE denotes HumanEval.
}
\label{tab:clad_params}
\begin{tabular}{@{}lcccc@{}}
\toprule
Model & GSM8K & MATH & HE & MBPP \\
\midrule
LLaDA-1.5  & 0.73 & 0.70 & 0.75 & 0.73 \\
LLaDA-Inst & 0.70 & 0.75 & 0.75 & 0.75 \\
Dream-Base & 0.75 & 0.70 & 0.73 & 0.75 \\
Dream-Inst & 0.75 & 0.70 & 0.71 & 0.70 \\
\bottomrule
\end{tabular}
\end{table}

\section{Additional Results on Other Models}
\label{app:additional_results}

We additionally report results on LLaDA-1.5 and Dream-v0-Base-7B in Table~\ref{tab:appendix_additional_models}. We follow the same evaluation setting and reporting format as in the main experiments. Acc. denotes task accuracy, TPS denotes tokens per second, and Speedup is measured relative to Vanilla decoding. Colored subscripts in the Acc. columns indicate the absolute accuracy change compared with Vanilla decoding. We further provide qualitative examples comparing \methodname{} with Vanilla decoding on LLaDA-8B-Instruct across the four evaluated tasks, showing the actual generated outputs of both decoding methods.

\begin{table*}[t]
\centering
\small
\setlength{\tabcolsep}{4pt}
\renewcommand{\arraystretch}{1.08}
\caption{Additional results on reasoning and code generation benchmarks with LLaDA-1.5 and Dream-v0-Base-7B. Acc. denotes task accuracy, TPS denotes tokens per second, and Speedup is measured relative to Vanilla top-1 decoding under the same setting. Colored subscripts on Acc. report the absolute change relative to Vanilla. Higher values are better for all metrics, and the best TPS and Speedup are highlighted in bold.}
\label{tab:appendix_additional_models}
\resizebox{\textwidth}{!}{
\begin{tabular}{llcccccc}
\toprule
\multirow{2}{*}{Task} & \multirow{2}{*}{Method}
& \multicolumn{3}{c}{LLaDA-1.5}
& \multicolumn{3}{c}{Dream-v0-Base-7B} \\
\cmidrule(lr){3-5} \cmidrule(lr){6-8}
& & Acc.$\uparrow$ & TPS$\uparrow$ & Speedup$\uparrow$
& Acc.$\uparrow$ & TPS$\uparrow$ & Speedup$\uparrow$ \\
\midrule

\multirow{6}{*}{GSM8K (5-shot)}
& Vanilla   & 81.12              & 4.90  & 1.00$\times$ & 75.13              & 7.16  & 1.00$\times$ \\
& Fast-dLLM & \accup{81.35}{0.23} & 16.61 & 3.39$\times$ & \accdown{74.15}{0.98} & 12.04 & 1.68$\times$ \\
& KLASS     & \accup{81.50}{0.38} & 4.77  & 0.97$\times$ & \accdown{72.40}{2.73} & 7.10  & 0.99$\times$ \\
& DAPD      & \accup{81.20}{0.08} & 10.86 & 2.22$\times$ & \accdown{71.95}{3.18} & 10.62 & 1.48$\times$ \\
& DAWN      & \accsame{81.12}     & 22.63 & 4.62$\times$ & \accdown{72.71}{2.42} & \textbf{13.39} & \textbf{1.87$\times$} \\
& \methodname{} (ours)      & \accup{81.65}{0.53} & \textbf{23.07} & \textbf{4.71$\times$} & \accdown{73.54}{1.59} & 12.67 & 1.77$\times$ \\
\midrule

\multirow{6}{*}{MATH (4-shot)}
& Vanilla   & 33.84              & 7.27  & 1.00$\times$ & 34.56              & 9.01  & 1.00$\times$ \\
& Fast-dLLM & \accdown{33.70}{0.14} & 19.35 & 2.66$\times$ & \accup{34.90}{0.34} & 21.76 & 2.42$\times$ \\
& KLASS     & \accdown{33.50}{0.34} & 6.97  & 0.96$\times$ & \accdown{30.70}{3.86} & 7.89  & 0.88$\times$ \\
& DAPD      & \accdown{33.04}{0.80} & 14.63 & 2.01$\times$ & \accdown{31.10}{3.46} & 18.16 & 2.02$\times$ \\
& DAWN      & \accdown{32.10}{1.74} & 25.03 & 3.44$\times$ & \accup{34.58}{0.02} & 25.68 & 2.85$\times$ \\
& \methodname{} (ours)       & \accdown{32.72}{1.12} & \textbf{26.78} & \textbf{3.68$\times$} & \accdown{33.98}{0.58} & \textbf{27.47} & \textbf{3.05$\times$} \\
\midrule

\multirow{6}{*}{MBPP (3-shot)}
& Vanilla   & 38.60              & 1.99  & 1.00$\times$ & 52.80              & 9.18  & 1.00$\times$ \\
& Fast-dLLM & \accdown{38.00}{0.60} & 11.78 & 5.92$\times$ & \accup{53.20}{0.40} & 25.35 & 2.76$\times$ \\
& KLASS     & \accdown{25.80}{12.80} & 1.45 & 0.73$\times$ & \accup{56.43}{3.63} & 8.12  & 0.88$\times$ \\
& DAPD      & \accsame{38.60}     & 3.89  & 1.95$\times$ & \accdown{49.20}{3.60} & 17.02 & 1.85$\times$ \\
& DAWN      & \accdown{37.40}{1.20} & 16.27 & 8.18$\times$ & \accup{53.73}{0.93} & \textbf{33.20} & \textbf{3.62$\times$} \\
& \methodname{} (ours)       & \accup{38.80}{0.20} & \textbf{16.86} & \textbf{8.47$\times$} & \accup{54.20}{1.40} & 29.06 & 3.17$\times$ \\
\midrule

\multirow{6}{*}{HumanEval (0-shot)}
& Vanilla   & 42.68              & 5.77  & 1.00$\times$ & 39.02              & 16.10 & 1.00$\times$ \\
& Fast-dLLM & \accsame{42.68}     & 15.00 & 2.60$\times$ & \accup{43.29}{4.27} & 30.18 & 1.87$\times$ \\
& KLASS     & \accsame{42.68}     & 5.46  & 0.95$\times$ & \accup{43.90}{4.88} & 16.98 & 1.05$\times$ \\
& DAPD      & \accdown{41.46}{1.22} & 10.83 & 1.88$\times$ & \accup{39.63}{0.61} & 24.73 & 1.54$\times$ \\
& DAWN      & \accdown{41.46}{1.22} & 19.15 & 3.32$\times$ & \accup{42.68}{3.66} & \textbf{33.68} & \textbf{2.09$\times$} \\
& \methodname{} (ours)       & \accsame{42.68}     & \textbf{20.71} & \textbf{3.59$\times$} & \accup{39.63}{0.61} & 33.31 & 2.07$\times$ \\
\bottomrule
\end{tabular}
}
\end{table*}


\begin{table*}[t]
\centering
\caption{Example generation output of LLaDA-8B-Instruct with Vanilla decoding on GSM8K (5-shot)}
\label{tab:gsm8k_5shot_llada_without_clad_case}
\renewcommand{\arraystretch}{1.25}
\begin{tabular}{p{0.18\textwidth} p{0.72\textwidth}}
\hline
\textbf{Item} & \textbf{Content} \\
\hline
\textbf{Question} &
Janet’s ducks lay 16 eggs per day. She eats three for breakfast every morning and bakes muffins for her friends every day with four. She sells the remainder at the farmers' market daily for \$2 per fresh duck egg. How much in dollars does she make every day at the farmers' market? \\
\hline
\textbf{Output} &
To determine how much Janet makes every day at the farmers' market, we need to follow these steps:

1. Calculate the total number of eggs laid by Janet's ducks per day.

2. Subtract the number of eggs she eats for breakfast and bakes for her friends.

3. Calculate the number of eggs she sells.

4. Multiply the number of eggs sold by the price per egg to find her daily earnings.

Janet's ducks lay 16 eggs per day. She eats 3 eggs for breakfast and bakes 4 eggs for her friends, so she eats a total of 3 + 4 = 7 eggs per day. The number of eggs she sells is the total number of eggs laid minus the number of eggs she eats, which is 16 - 7 = 9 eggs per day. She sells each egg for \$2, so her daily earnings are 9 eggs * \$2/egg = \$18.

Therefore, Janet makes \$18 every day at the farmers' market. \\
\hline
\textbf{Generation Time} & 48.92 s \\
\hline
\end{tabular}
\end{table*}

\begin{table*}[!t]
\centering
\caption{Example generation output of LLaDA-8B-Instruct with \methodname{} on GSM8K (5-shot)}
\label{tab:gsm8k_5shot_llada_with_clad_case}
\renewcommand{\arraystretch}{1.25}
\begin{tabular}{p{0.18\textwidth} p{0.72\textwidth}}
\hline
\textbf{Item} & \textbf{Content} \\
\hline
\textbf{Question} &
Janet’s ducks lay 16 eggs per day. She eats three for breakfast every morning and bakes muffins for her friends every day with four. She sells the remainder at the farmers' market daily for \$2 per fresh duck egg. How much in dollars does she make every day at the farmers' market? \\
\hline
\textbf{Output} &
To determine how much Janet makes every day at the farmers' market, we need to follow these steps:

1. Calculate the total number of eggs laid by Janet's ducks per day.

2. Subtract the number of eggs Janet eats for breakfast and bakes for her friends.

3. Calculate the number of eggs left for sale.

4. Multiply the number of eggs left for sale by the price per egg to find the daily earnings.

1. Janet's ducks lay 16 eggs per day.

2. She eats 3 eggs for breakfast and bakes 4 eggs for her friends, so she uses a total of 3 + 4 = 7 eggs per day.

3. The number of eggs left for sale is the total number of eggs laid minus the number of eggs used, which is 16 - 7 = 9 eggs per day.

4. She sells each egg for \$2, so her daily earnings are the number of eggs left for sale multiplied by the price per egg, which is 9 eggs * \$2/egg = \$18 per day.

Therefore, Janet makes \$18 every day at the farmers' market.

\#\#\#\# 18 \\
\hline
\textbf{Generation Time} & 13.07 s \\
\hline
\end{tabular}
\end{table*}

\begin{table*}[!t]
\centering
\caption{Example generation output of LLaDA-8B-Instruct with Vanilla decoding on MATH (4-shot)}
\label{tab:math_4shot_llada_without_clad_case}
\renewcommand{\arraystretch}{1.25}
\begin{tabular}{p{0.18\textwidth} p{0.72\textwidth}}
\hline
\textbf{Item} & \textbf{Content} \\
\hline
\textbf{Question} &
How many vertical asymptotes does the graph of $y=\frac{2}{x^2+x-6}$ have? \\
\hline
\textbf{Output} &
To determine the number of vertical asymptotes of the function \( y = \frac{2}{x^2 + x - 6} \), we need to identify the values of \( x \) that make the denominator zero, as these are the points where the function is undefined and may have vertical asymptotes.

First, we factor the denominator \( x^2 + x - 6 \):

\[
x^2 + x - 6 = (x + 3)(x - 2)
\]

Next, we set each factor equal to zero and solve for \( x \):

\[
x + 3 = 0 \implies x = -3
\]

\[
x - 2 = 0 \implies x = 2
\]

Thus, the values of \( x \) that make the denominator zero are \( x = -3 \) and \( x = 2 \). These are the points where the function has vertical asymptotes.

Therefore, the graph of \( y = \frac{2}{x^2 + x - 6} \) has two vertical asymptotes.

The final answer is:

\[
\boxed{2}
\] \\
\hline
\textbf{Generation Time} & 19.54 s \\
\hline
\end{tabular}
\end{table*}

\begin{table*}[!t]
\centering
\caption{Example generation output of LLaDA-8B-Instruct with \methodname{} on MATH (4-shot)}
\label{tab:math_4shot_llada_with_clad_case}
\renewcommand{\arraystretch}{1.25}
\begin{tabular}{p{0.18\textwidth} p{0.72\textwidth}}
\hline
\textbf{Item} & \textbf{Content} \\
\hline
\textbf{Question} &
How many vertical asymptotes does the graph of $y=\frac{2}{x^2+x-6}$ have? \\
\hline
\textbf{Output} &
To determine the number of vertical asymptotes of the function \( y = \frac{2}{x^2 + x - 6} \), we need to find the values of \( x \) that make the denominator zero, as vertical asymptotes occur where the function is undefined.

First, we set the denominator equal to zero and solve for \( x \):
\[
x^2 + x - 6 = 0
\]

Next, we factor the quadratic equation:
\[
(x + 3)(x - 2) = 0
\]

Setting each factor equal to zero gives us the solutions:
\[
x + 3 = 0 \quad \text{and} \quad x - 2 = 0
\]
\[
x = -3 \quad \text{and} \quad x = 2
\]

These values of \( x \) make the denominator zero, and thus the function \( y = \frac{2}{x^2 + x - 6} \) has vertical asymptotes at \( x = -3 \) and \( x = 2 \).

Therefore, the number of vertical asymptotes is:
\[
\boxed{2}
\] \\
\hline
\textbf{Generation Time} & 4.91 s \\
\hline
\end{tabular}
\end{table*}

\clearpage

\begin{table*}[!t]
\centering
\caption{Example generation output of LLaDA-8B-Instruct with Vanilla decoding on MBPP (3-shot)}
\label{tab:mbpp_3shot_llada_without_clad_case}
\renewcommand{\arraystretch}{1.25}
\begin{tabular}{p{0.18\textwidth} p{0.72\textwidth}}
\hline
\textbf{Item} & \textbf{Content} \\
\hline
\textbf{Task} &
Write a function to find whether a given array of integers contains any duplicate element. \\
\hline
\textbf{Output} &
\begin{minipage}[t]{0.72\textwidth}
\ttfamily\small
def test\_duplicate(arr):\par
\quad seen = set()\par
\quad for num in arr:\par
\quad\quad if num in seen:\par
\quad\quad\quad return True\par
\quad\quad seen.add(num)\par
\quad return False
\par\vspace{0.45em}
\end{minipage}
\\
\hline
\textbf{Reference Tests} &
\begin{minipage}[t]{0.72\textwidth}
\ttfamily\small
assert test\_duplicate(([1,2,3,4,5]))==False\par
assert test\_duplicate(([1,2,3,4, 4]))==True\par
assert test\_duplicate([1,1,2,2,3,3,4,4,5])==True
\par\vspace{0.45em}
\end{minipage}
\\
\hline
\textbf{Evaluation Result} &
Passed, pass@1 = 1.0 \\
\hline
\textbf{Generation Time} & 28.21 s \\
\hline
\end{tabular}
\end{table*}

\begin{table*}[!t]
\centering
\caption{Example generation output of LLaDA-8B-Instruct with \methodname{} on MBPP (3-shot)}
\label{tab:mbpp_3shot_llada_with_clad_case}
\renewcommand{\arraystretch}{1.25}
\begin{tabular}{p{0.18\textwidth} p{0.72\textwidth}}
\hline
\textbf{Item} & \textbf{Content} \\
\hline
\textbf{Task} &
Write a function to find whether a given array of integers contains any duplicate element. \\
\hline
\textbf{Output} &
\begin{minipage}[t]{0.72\textwidth}
\ttfamily\small
def test\_duplicate(arr):\par
\quad seen = set()\par
\quad for num in arr:\par
\quad\quad if num in seen:\par
\quad\quad\quad return True\par
\quad\quad seen.add(num)\par
\quad return False
\par\vspace{0.45em}
\end{minipage}
\\
\hline
\textbf{Reference Tests} &
\begin{minipage}[t]{0.72\textwidth}
\ttfamily\small
assert test\_duplicate(([1,2,3,4,5]))==False\par
assert test\_duplicate(([1,2,3,4, 4]))==True\par
assert test\_duplicate([1,1,2,2,3,3,4,4,5])==True
\par\vspace{0.45em}
\end{minipage}
\\
\hline
\textbf{Evaluation Result} &
Passed, pass@1 = 1.0 \\
\hline
\textbf{Generation Time} & 2.23 s \\
\hline
\end{tabular}
\end{table*}

\begin{table*}[!t]
\centering
\caption{Example generation output of LLaDA-8B-Instruct with Vanilla decoding on HumanEval (0-shot)}
\label{tab:humaneval_0shot_llada_without_clad_case}
\begingroup
\setlength{\tabcolsep}{4pt}
\renewcommand{\arraystretch}{1.18}
\begin{tabular}{
@{}>{\raggedright\arraybackslash}p{0.24\textwidth}
>{\raggedright\arraybackslash}p{0.72\textwidth}@{}
}
\hline
\rowhead{Item} & \textbf{Content} \\
\hline
\rowhead{Task} &
Calculate the mean absolute deviation of a given list of numbers around its mean. \\
\hline
\rowhead{Prompt} &
\begin{minipage}[t]{\linewidth}
\vspace{0.2em}
\begin{CodeBlock}
from typing import List

def mean_absolute_deviation(numbers: List[float]) -> float:
    """ For a given list of input numbers, calculate Mean Absolute Deviation
    around the mean of this dataset.
    Mean Absolute Deviation is the average absolute difference between each
    element and a centerpoint (mean in this case):
    MAD = average | x - x_mean |
    >>> mean_absolute_deviation([1.0, 2.0, 3.0, 4.0])
    1.0
    """
\end{CodeBlock}
\vspace{0.2em}
\end{minipage}
\\
\hline
\rowhead{Output} &
\begin{minipage}[t]{\linewidth}
\vspace{0.2em}
\begin{CodeBlock}
from typing import List

def mean_absolute_deviation(numbers: List[float]) -> float:
    """ For a given list of input numbers, calculate Mean Absolute Deviation
    around the mean of this dataset.
    Mean Absolute Deviation is the average absolute difference between each
    element and a centerpoint (mean in this case):
    MAD = average | x - x_mean |
    >>> mean_absolute_deviation([1.0, 2.0, 3.0, 4.0])
    1.0
    """
    if not numbers:
        return 0.0

    mean_value = sum(numbers) / len(numbers)
    absolute_deviations = [abs(x - mean_value) for x in numbers]
    mad = sum(absolute_deviations) / len(absolute_deviations)
    return mad

if __name__ == "__main__":
    import doctest
    doctest.testmod()
\end{CodeBlock}
\vspace{0.2em}
\end{minipage}
\\
\hline
\rowhead{Reference Tests} &
\begin{minipage}[t]{\linewidth}
\vspace{0.2em}
\begin{CodeBlock}
assert abs(candidate([1.0, 2.0, 3.0]) - 2.0/3.0) < 1e-6
assert abs(candidate([1.0, 2.0, 3.0, 4.0]) - 1.0) < 1e-6
assert abs(candidate([1.0, 2.0, 3.0, 4.0, 5.0]) - 6.0/5.0) < 1e-6
\end{CodeBlock}
\vspace{0.2em}
\end{minipage}
\\
\hline
\rowhead{Evaluation Result} &
Passed, pass@1 = 1.0 \\
\hline
\rowhead{Generation Time} &
15.10 s \\
\hline
\end{tabular}
\endgroup
\end{table*}

\begin{table*}[!t]
\centering
\caption{Example generation output of LLaDA-8B-Instruct with \methodname{} on HumanEval (0-shot)}
\label{tab:humaneval_0shot_llada_with_clad_case}
\begingroup
\setlength{\tabcolsep}{4pt}
\renewcommand{\arraystretch}{1.18}
\begin{tabular}{
@{}>{\raggedright\arraybackslash}p{0.24\textwidth}
>{\raggedright\arraybackslash}p{0.72\textwidth}@{}
}
\hline
\rowhead{Item} & \textbf{Content} \\
\hline
\rowhead{Task} &
Calculate the mean absolute deviation of a given list of numbers around its mean. \\
\hline
\rowhead{Prompt} &
\begin{minipage}[t]{\linewidth}
\vspace{0.2em}
\begin{CodeBlock}
from typing import List

def mean_absolute_deviation(numbers: List[float]) -> float:
    """ For a given list of input numbers, calculate Mean Absolute Deviation
    around the mean of this dataset.
    Mean Absolute Deviation is the average absolute difference between each
    element and a centerpoint (mean in this case):
    MAD = average | x - x_mean |
    >>> mean_absolute_deviation([1.0, 2.0, 3.0, 4.0])
    1.0
    """
\end{CodeBlock}
\vspace{0.2em}
\end{minipage}
\\
\hline
\rowhead{Output} &
\begin{minipage}[t]{\linewidth}
\vspace{0.2em}
\begin{CodeBlock}
from typing import List

def mean_absolute_deviation(numbers: List[float]) -> float:
    """ For a given list of input numbers, calculate Mean Absolute Deviation
    around the mean of this dataset.
    Mean Absolute Deviation is the average absolute difference between each
    element and a centerpoint (mean in this case):
    MAD = average | x - x_mean |
    >>> mean_absolute_deviation([1.0, 2.0, 3.0, 4.0])
    1.0
    """
    if not numbers:
        return 0.0

    mean_value = sum(numbers) / len(numbers)
    absolute_deviations = [abs(x - mean_value) for x in numbers]
    mad = sum(absolute_deviations) / len(numbers)
    return mad

if __name__ == "__main__":
    import doctest
    doctest.testmod()
\end{CodeBlock}
\vspace{0.2em}
\end{minipage}
\\
\hline
\rowhead{Reference Tests} &
\begin{minipage}[t]{\linewidth}
\vspace{0.2em}
\begin{CodeBlock}
assert abs(candidate([1.0, 2.0, 3.0]) - 2.0/3.0) < 1e-6
assert abs(candidate([1.0, 2.0, 3.0, 4.0]) - 1.0) < 1e-6
assert abs(candidate([1.0, 2.0, 3.0, 4.0, 5.0]) - 6.0/5.0) < 1e-6
\end{CodeBlock}
\vspace{0.2em}
\end{minipage}
\\
\hline
\rowhead{Evaluation Result} &
Passed, pass@1 = 1.0 \\
\hline
\rowhead{Generation Time} &
2.69 s \\
\hline
\end{tabular}
\endgroup
\end{table*}

\end{document}